# Identification of non-linear behavior models with restricted or redundant data

**S. Carbillet, V. Guicheret-Retel, F. Trivaudey, F. Richard, M.L. Boubakar**

Univ. Bourgogne Franche-Comté – FEMTO-ST Institute, CNRS/UFC/ENSMM/UTBM,

Department of Applied Mechanics, Besançon, France

Corresponding author: Violaine Guicheret-Retel, violaine.retel@univ-fcomte.fr, 24, rue de l'Epitaphe, 25000 Besançon, France, tel : +33 3 63 08 25 60, fax : +33 3 81 66 67 00.

**Abstract**

This study presents a new strategy for the identification of material parameters in the case of restricted or redundant data, based on a hybrid approach combining a genetic algorithm and the Levenberg-Marquardt method. The proposed methodology consists essentially in a statistically based topological analysis of the search domain, after this one has been reduced by the analysis of the parameters ranges. This is used to identify the parameters of a model representing the behavior of damaged elastic, visco-elastic, plastic and visco-plastic composite laminates. Optimization of the experimental tests on tubular samples leads to the selective identification of these parameters.

**1- Introduction**

The increasing complexity of material behavior models has led to an increasing number of parameters requiring identification. This is the case for the model presented in [5,6], describing the damaged elastic, visco-elastic, plastic, and visco-plastic behavior of composites laminates, made with an organic matrix and reinforced with long fibers. The 21 parameters of the layer model are determined from experimental data. This is an inverse problem, which can be reduced to one of parameter optimization. Nevertheless, the volume of experimental data available for the identification of these parameters is insufficient to obtain a well-conditioned problem. This is generally the case with expensive materials or when experimental tests are difficult to perform.

The aim of the present study is to propose a parameter identification strategy for the particular case of a restricted quantity of experimental data. Classical local algorithms are no longer efficient in the case of a multi-modal minimization problem [27]: multiple parameter sets often exist, all of which lead to quasi-similar responses. Under such conditions, there is no discriminating solution.

Research in the geotechnical field [18,25], in which the error function topology is often complex and there is no unique solution, shows that the genetic algorithm is particularly suitable for such cases. One advantage of genetic algorithms is that only the function to be maximized, but not its derivatives or other quantities, needs to be evaluated. This property allows for the exploration of large domains. Genetic algorithms belong to the family of "metaheuristic" methods, which include artificial neuronal networks [7, 28], the simulated annealing method [11,14,16-17], the tabu search method [29], and ant colony research [2]. One advantage of such methods is their ability to proceed to a multi-objective optimization: the best individuals can be evaluated with several objective functions.

Although these algorithms are not efficient in finding the ideal solution, they are very useful for the detection of the zone of interest. For this reason, they are often coupled with local solving methods [23-24]. The present study presents a new identification strategy based on a statistical analysis and the use of a hybrid approach, combining a genetic algorithm with the local Levenberg-Marquardt method. The genetic algorithm is used to prospect for the solution area, avoiding local optima, in the search domain which can be relatively large in size. The local method is then used to accurately localize the optimal solution.

In section 2 the hybrid approach and our original identification methodology are described. The search domain is initially reduced in size by means of statistical analysis of the distribution of each parameter to be identified. The hybrid algorithm is then used, with an original strategy involving a statistical approach to the topological analysis of the search domain. The constitutive equations of the model [5,6], as well as the 21 parameters to be identified, are described in section 3. Section 4 presents the different tests used for this identification, for which the experimental data was obtained by performing multi-axial tests on composites tubes. These tests included tension, pure internal pressure, internal pressure with end-effect tests, and creep tests under tension. The identification process and its corresponding results are then presented in section 5. By optimizing the experimental tests it was possible to discriminate between the different components of the behavior law, thus allowing the relevant parameters to be selectively identified. The identified behavior is validated through the use of specific tests. Finally, the results and methodology are discussed in the last section of this paper.

**2- Materials and methods**

The identification of material parameters can be considered as an inverse formulation [8,12,22]. In this context, the optimization process involves running a numerical simulation of the performed experiment, whilst trying to adjust the $\alpha$ material parameters $\boldsymbol{\theta} = {}^T[\theta_1, ..., \theta_\alpha]$, to obtain simulation and experimental results as closed as possible. The material parameters are optimized by minimizing the differences between the experimental results $m(t)$ and the numerical results $h(\boldsymbol{\theta}, t)$, for any value of the time variable $t$ belonging to the period of the experimental test $[0, t_N]$. In the present study, the experimental results produced by only one test, and one sensor, are considered. The differences (or residuals) between the experimental and numerical results can be defined as an implicit function of $\boldsymbol{\theta}$:

$$j(\boldsymbol{\theta}, t) = m(t) - h(\boldsymbol{\theta}, t) \qquad (1)$$

The solution to the identification problem then consists in the minimization of a scalar function $f(\boldsymbol{\theta})$, expressed as:

$$f(\boldsymbol{\theta}) = \frac{1}{2N} \sum_{i=1}^{N} V_i j^2(\boldsymbol{\theta}, t_i) \qquad (2)$$

where $N$ is the number of acquisitions, $t_i$ is the instant in time corresponding to the $i^{th}$ acquisition, and $V_i$ is a weighting factor associated with each data point $i$. This factor allows the uncertain nature of some of the observed variables to be taken into account and make the function $f$ dimensionless (essential if several sensors are used). The weighting coefficients proposed here are:

$$V_i = V = \frac{1}{\chi} \text{ with } \chi = \left[\max_{t_i}(|m(t_i)|)\right]^2 \qquad (3)$$

This leads to:

$$f(\boldsymbol{\theta}) = \frac{1}{2N} \sum_{i=1}^{N} \frac{1}{\chi} j^2(\boldsymbol{\theta}, t_i) \qquad (4)$$

This choice allows the definition of a higher weight for the variables experimentally measured with a higher precision [1].

The cost function given in Eq. 4 is defined for a single experimental test, the results of which are provided by a given sensor. For the case of several experimental tests ($k$) and multiple sensors ($j$), the following cost function can be written:

$$f(\theta) = \frac{1}{N_{test}} \sum_{k=1}^{N_{test}} \left[ \frac{1}{N_S} \sum_{j=1}^{N_S} \left( \frac{1}{2N} \sum_{i=1}^{N} \frac{1}{\chi_{(k,j)}} j_{(k,j)}^2(\theta, t_i) \right) \right]$$

hence

$$f(\theta) = \frac{1}{2N_{test}N_S N} \sum_{k=1}^{N_{test}} \sum_{j=1}^{N_S^{(k)}} \left( \frac{1}{\chi_{(k,j)}} \sum_{i=1}^{N^{(k,j)}} \left( m_{(k,j)}(t_i) - h_{(k,j)}(\theta, t_i) \right)^2 \right)$$

where $N_{test}$ is the number of tests, $N_S^{(k)}$ is the number of sensors for the $(k)^{th}$ test, and $N^{(k,j)}$ is the number of temporal acquisitions corresponding to the $(j)^{th}$ sensor, and the $(k)^{th}$ test. One weighting coefficient is considered for each sensor, during each test.

The minimization problem can be formally written as:

$$\theta^* = \underset{\theta \in [\theta^-, \theta^+]}{\mathrm{argmin}}\, f(\theta) \tag{5}$$

where $\theta^*$ is the solution for the identification problem, and $\theta^-$ and $\theta^+$ are respectively the minimum and maximum limits of the material parameters.

In this study, it was decided to use a hybrid algorithm for the identification process, combining a genetic algorithm with the Levenberg-Marquardt method. This algorithm is illustrated in Fig. 1 and is described in detail in section 2.1. In the following, the identification strategy is presented in the context of restricted or redundant data.

## 2.1- Hybrid algorithm

Genetic algorithms were introduced by Holland [15] in the seventies. They are based on the genetic evolutionary process of natural species first described by Darwin, in which only those individuals best adapted to the environment survive and have offspring (lineage). This type of evolutionary algorithm does not need to calculate the derivatives of the objective function, and is thus able to solve problems in which explicit formulations are not easy to determine, and for which classical optimization algorithms are inefficient. The solution is sought by maintaining a population of individuals, which represent the potential solutions. This population evolves through the action of probabilistic operators, representing the natural selection-reproduction-replacement cycle.

The initial population is obtained from a random draw of $N_{ind}$ individuals. Each of them is characterized by its genes, which are the $\alpha$ parameters to be identified. $N_{ind}$ is chosen to be equal to $10\alpha$. Each individual is evaluated and compared to the others by means of an adaptive function $f_a(\boldsymbol{\theta})$. The genetic algorithm tends to maximize this function. Since the identification process tends to minimize the objective function $f(\boldsymbol{\theta})$, the adaptive function is defined as the inverse of the objective function.

The genetic operators used to cause the population to evolve are selection, cross-over and mutation. The selection operator is designed to select the better individuals from a given generation (named the 'parents'), to create the next (new) generation (the 'children'). The selection of parents in the current population is very important, insofar as it determines the genetic content of the next population. This step must bring to optima, whilst maintaining a certain degree of diversity in the population. The selection method is a mode, which is proportional to the adaptive function value of each individual. The aim of the cross-over operator is to combine the genes of the parents to produce children. It allows the population diversity to be enriched, and investigates all possible solutions. The parents ($x_1$, $x_2$) are matched to form a couple, which produces a pair of children ($y_1$, $y_2$), such that the population always comprises the same number of individuals. The cross-over operator consists in a linear combination of parents (See Eq. 6), depending on the crossbreed probability, the value of which is commonly chosen to lie between 0.6 and 1. In this study, the value of 0.8 was assumed.

$$y_1 = ax_1 + (1-a)x_2 \quad \text{and} \quad y_2 = (1-a)x_1 + ax_2 \tag{6}$$

where *a* is a random number chosen between 0 and 1. The mutation operator acts randomly, and locally changes the genetic content of an individual. The role of this operator consists in preserving a certain degree of diversity in the population, to avoid premature convergence towards a local solution. The occurrence of mutations depends on the mutation probability which, in this study, was chosen to be equal to *2/N*, providing a good compromise between the convergence rate and a relatively broad exploration of the domain of potential solutions [13].

Among the various criteria which could be used to stop the algorithm, that defining a maximum number of generations was chosen in this study. The genetic algorithm was stopped when it had produced 30 successive generations.

One drawback of genetic algorithms is that it is difficult to adjust all of the parameters (crossbreed and mutation probabilities, size of the initial population, stop criterion, …) influencing the solution. For this reason it was decided to keep the simplest form of the algorithm, with a minimum number of adjustment parameters. Further details concerning the algorithm parameters can be found in [10]. Although the genetic algorithm used in this study is rather simplified, this is compensated for by the strategy presented in the following, together with the use of optimized tests for the generation of experimental data.

The genetic algorithm is coupled with a gradient-based algorithm, referred to as the Levenberg-Marquardt scheme [19-20]. Given a starting point $\theta^{(k)}$, a quadratic approximation to the objective function $f(\theta^{(k)})$ is determined in such a way as to match the values of the first and second derivatives at that point. Then, this approximate function, instead of the original objective function, is minimized. The approximate function minimizer is used as the starting point for the next step, and the procedure is repeated iteratively. If the objective function is quadratic, then the approximation is exact, and the method yields the true minimizer in one step. On the other hand, if the objective function is not quadratic, then the approximation will provide an estimate of the true minimizer. In the $k^{th}$ iteration of the optimization process, the step size $d\theta^{(k)} = \theta^{(k+1)} - \theta^{(k)}$ is determined by

$$d\theta^{(k)} = -\left[H^{(k)} + \lambda^{(k)} I\right]^{-1} g^{(k)} \tag{7}$$

where $g^{(k)} = g(\theta^{(k)})$ is the gradient of $f$ with respect to $\theta$ in $\theta^{(k)}$, and $H^{(k)} = H(\theta^{(k)})$, the Hessian matrix:

$$g^{(k)} = {}^T J^{(k)} f^{(k)} \text{ and } H^{(k)} = {}^T J^{(k)} J^{(k)}$$

where $J$ denotes the Jacobian matrix of the function $f$. This matrix is evaluated through the use of a numerical estimation based on finite differentiation:

$$J_{i\alpha}^{(k)} = \left.\frac{\partial f_i}{\partial \theta_\alpha}\right|_{\theta=\theta^{(k)}} \approx \frac{\Delta f_i^{(k)}}{\Delta \theta_\alpha} \quad i = 1, \dots, n \, ; \, \theta = 1, \dots, \alpha$$

In order to facilitate the choice of the damping factor $\lambda^{(k)}$, Eq. (7) is normalized before being solved.

$$d\underline{\theta}^{(k)} = -\left[\underline{H}^{(k)} + \lambda^{(k)} I\right]^{-1} \underline{g}^{(k)}$$

with $\underline{g}_i^{(k)} = \frac{g_i^{(k)}}{\sqrt{H_{ii}^{(k)}}}$ and $\underline{H}_{ij}^{(k)} = \frac{H_{ij}^{(k)}}{\sqrt{H_{ii}^{(k)} H_{jj}^{(k)}}}$

such that $d\boldsymbol{\theta}^{(k)}$ is given by $d\underline{\theta}_i^{(k)} = \frac{d\theta_i^{(k)}}{\sqrt{H_{ii}^{(k)}}}$

$\lambda^{(k)}$ is adapted dynamically according to a heuristic rule, whilst being restricted to the limits: $\boldsymbol{\theta}^- \leq \boldsymbol{\theta}^{(k)} \leq \boldsymbol{\theta}^+$ as defined in [26]. This algorithm was implemented in the MIC2M® [21] program.

The starting point of the Levenberg-Marquardt algorithm is defined by the best individual, noted as $\boldsymbol{\theta}^{(0)}$, extracted from the previous population produced by the genetic algorithm.

**2.2- Identification strategy with restricted or redundant data**

In genetic algorithms, the parameters (genes of the individuals) must be encoded. Although binary encoding is often used in combinatory problems [13], in the present context the problem is not combinatory, and since the sought parameters are real, real encoding is preferred. The advantage of real encoding is that it allows the parameters to be varied continuously within their defined domain, when they are subjected to stochastic operators. This property is used to evaluate the spatial distribution of the parameters, and then to reduce the search domain. This is achieved by drawing a map indicating the individual density of each parameter.

A first method was proposed in [4], involving the discretization of a given parameter's initial domain into a set of sub-domains. The density of each sub-domain is then evaluated by counting the number of occurrences of the parameter in this sub-domain.

This process is used to produce a histogram, which can be used to reduce the size of the search domain.

A different method was proposed in [3], in which, for each parameter, a scatter diagram is made, showing the value of the adaptive function as a function of the parameter's value. This is achieved by using the genetic algorithm with a mutation probability equal to 1, which allows a uniform search to be made over the entire studied domain. A threshold value is chosen for the adaptive function, such that only those parameter values greater than the threshold are kept and plotted. Several cases are encountered, as shown in Fig. 2. The first case corresponds to that of a parameter with a uniform distribution, indicating that the parameter has very little influence with respect to the adaptive function. In such a case, the studied domain cannot be reduced. In the second case, several peaks can be seen, each of which corresponds to a possible local solution. The last case indicates the existence of a dominant region, which necessarily contains the optimal solution. These distribution representations can be used to reduce the search domain. After several runs of the genetic algorithm, regions of no interest can appear. If they are located near to the edge of the domain, they can be removed. Through this reduction in its size, the final search domain allows the efficiency of the genetic algorithm to be improved.

This second method for search domain reducing was used in our study, as it is considered to be more efficient when the physical space of the parameters is a priori known.

Genetic algorithms are very useful when searching for the approximate solution to a problem, or determining the coordinates of local optima. Nevertheless, in the context of the present study the chosen objective function was rather simple, and a restricted volume of data was available as a consequence of the small number of experimental tests. The identification problem is thus multimodal, with no discriminating solutions. To overcome this difficulty, an original strategy was designed, allowing a satisfactory solution to be obtained. This involved the implementation of statistical analysis to perform a topological evaluation of the restricted search domain: the hybrid algorithm was run several times and the different solutions were analyzed.

The proposed methodology (described in Fig. 3) consists so firstly in choosing the initial search domain, by taking into account the physical characteristics of the studied material. Then the method of domain reduction is applied (by launching ten times the genetic algorithm with a mutation probability of 1). The following step consists in running ten times the hybrid algorithm on the reduced search domain and analyzing the obtained solutions. If a unique solution is obtained, this is the optimal solution. If different solutions are obtained, then a topological analysis is done by calculating the standard-deviation of the parameters and the response dispersion. Several cases can be found:

- if the response dispersion is large, then the identification can be refined by reducing again the search domain (the new domain is then defined by the extreme values of each parameter obtained). Then, the process is repeated by launching ten times the hybrid algorithm on this new domain and doing again the topological analysis of the solutions,
- if the response dispersion is low, then all set of parameters obtained is an acceptable solution. The standard-deviations of the parameters give an idea of the topology of the solution that can be rather like a basin, or rather like distinct peaks, or potentially a mix of them. This analysis allows besides to define the distribution of each parameter if needed for a stochastic approach.

## 3- Single ply constitutive laws

The identification process was applied, in order to identify the parameters of the model proposed in [5-6]. This phenomenological model was developed for the study of laminated composites made from unidirectional layers of a polymer matrix reinforced with long fibers. It is based on a meso-macro approach, in the thermodynamics framework of irreversible processes. It was proposed to use damaged elastic, visco-elastic, plastic, and visco-plastic behaviors of the single ply, characterized by 21 parameters. The full model is described in [5-6]. Table 1 summarizes the 21 parameters of the model to be identified.

## 4- Identification tests

The material parameters described in section 3 were identified by using the results of multi-axial tests. The samples were cylindrical tubes [+55, -55]$_6$, 280 mm in length, 60 mm in diameter, with a thickness of 2 mm, made from a pre-impregnated Vetrotex RC 10800 P9HTC2 composite, with R glass fibers and a Ciba-Geigy epoxy resin. They were produced by filament winding.

The tests were optimized in order to allow the different components of the material behavior law to be distinguished. Progressive and repeated tension, pure internal pressure, and internal pressure with end effects loadings, allowed the time-independent components of the material's behavior law to be identified (See Fig. 4). The tension loading consists in an axial stress progressive and repeated loading at a rate of 4 MPa/s with steps at 24, 34, 44, 54 and 60 MPa. The pure internal pressure loading corresponds to a hoop stress progressive and repeated loading at a rate of 8 MPa/s with steps at 80, 120, 160 and 200 MPa. Last, the internal pressure with end effects consists in an axial stress progressive and repeated loading at a rate of 4 MPa/s with steps at 40, 80, 100 and 120 MPa and a hoop stress progressive and repeated loading at a rate of 8 MPa/s with steps at 80, 160, 200 and 240 MPa.

A creep test under tension was used to identify the material's behavior law time-dependent component. The latter test involved the use of a 15-day plateau during which an axial stress of 20, 30, 0 and 40 MPa was applied. This type of test allows the various characteristics of the material's behavior to be activated separately, and the identification process to be broken down into several steps. During the tension test, the material's behavior is initially elastic, and then becomes non-linear, leading to induced residual strain. The pure internal pressure test allows the material to be loaded with a very high level of shear stress. The internal pressure with end effect loading was well tolerated by the samples, making this test very useful for the identification of the various damage parameters. Finally, the creep test activates the composite's visco-elastic behavior, without producing any damage.

For these tests, strain gauge rosettes have been used to obtain both radial and hoop strains on the exterior surface of the samples. Two rosettes are placed in the useful area on opposite faces of the specimen, in order to have mean values of the measured strain. Besides, a stress sensor or a pressure sensor (depending on the performed test) is used to evaluate the real loading the sample is subjected to. This information is needed to perform the finite element simulation, whose results are compared to experimental ones.

## 5- Results and Discussion

As described in the previous section, the elastic parameters could be identified separately from the non-elastic parameters, through the use of the first loading step of the tension test. The visco-elastic parameters are then given by the creep test. Finally, the remaining parameters are identified by analyzing the data from all of the tests. The following section describes the results of the identification, whereas section 5.2 deals with the validation of these results.

The CASTEM® and MIC2M® [21] computer codes were used for this identification process: the former gives the modeling results, whereas the latter is used for the statistical analysis of the data.

### 5.1- Identification results

The elastic parameters which can be identified are $E_1$, $E_2$, $\nu_{12}$, $G_{12}$ and $G_{23}$, whereas the shear modulus $G_{23}$ cannot be identified with the available tests. It was therefore obtained through the use of an asymptotic homogenization technique [9]. To begin the identification process, each parameter's range was defined and its distribution evaluated, in order to restrict the size of the search domain (See Fig. 5). $E_1$, $E_2$ and $G_{12}$ distributions show a preferential zone where the objective function has a low value, hence the size of the search domain used for these three parameters could be reduced, and centered on the peak value of each parameter. In the case of $\nu_{12}$, its distribution was found to have a plateau, indicating that this parameter has no influence on the adaptive function. Nevertheless, various physical considerations could be used to reduce its search domain. Table 2 lists the initial and reduced search ranges used for each parameter.

The second step consisted in running the hybrid algorithm, in order to evaluate the optimal solution. Ten draws were used, with an initial population of 40 members, and 30 generations for each draw. The elastic parameters are identified here, from the three tests presented in section 4 (tension, pure internal pressure and internal pressure with end effect). The results shown in Table 3 emphasize the multi-modal characteristics of the identification, when a small quantity of data is available: several distinct, realistic solutions are obtained, producing very similar objective function values. Each parameter can be seen to have a significant standard deviation, such that many different discriminating solutions could be found. This outcome results from a lack of experimental data (only three tests) available for the identification of four parameters. In order to compensate for this somewhat inadequate volume of experimental data, and to increase the robustness of the identification, tension/internal pressure like tests were added. Three axial-to-hoop stress ratios were used: 2/3, 1/3 and 1/6, and the hybrid algorithm was then run again 10 times. The results obtained from this analysis are summarized in Table 4. The response dispersions are seen to be very low, indicating convergence towards a global optimum. The finally retained set of parameters, listed in Table 5, is that having the lowest objective function value. These results show that the use of redundant tests allows the optimal solution to be discriminated. However, it can also be seen that the results of Table 4 lie within the dispersion of the results of Table 3, showing that, even when a small volume of experimental results is available, the domain of the optimal solution is correctly identified by the hybrid algorithm.

To determine the corresponding values of the remaining parameters, a step-by-step process was applied, using different parts of the experimental results. Firstly, the visco-elastic parameters were identified from the creep test results. Only the first plateau of the curve was used, and it was assumed that neither damage nor plasticity occurred. This assumption was verified at a later stage. The parameters corresponding to the damage and plasticity components were then identified from the results of the three progressive, repeated loading tests. For this aspect of the identification, the previously identified viscoelastic parameters were used. Finally, the visco-plastic parameters were identified from the two first plateaux of the creep test, by using the previously identified parameters corresponding to the other components of the behavior law.

The full set of results is presented in Table 5.

The experimental and simulation results are compared in Figs. 6 and 7. In Figs. 6a and 6c it can be seen that the stress levels are correctly modeled, and the non-linearity of the curve during loading is well respected. Nevertheless, Fig. 6a indicates a slight overestimation of residual stress, and Fig. 6c indicates an underestimation of residual stress during unloading. Figs. 6b and 7 show that a good agreement is found between the results produced by the model, using the identified parameters and the experimental measurements.

### 5.2- Validation

In order to validate the parameter identification described above, two kinds of test, different to those used in the identification process, were performed. The first of these consisted in a single, load-unload cycle of pure internal pressure on a tubular specimen. The specimen's geometry and material were the same as those described in section 4. The load rate was 10 MPa.s$^{-1}$ and the maximum applied pressure was 200 MPa. The results are presented in Fig. 8, showing a good correlation between the simulation and the experiment results, with a very slight under-estimation of the maximum strain and a slight over-estimation of the residual strain.

The second validation test involved the bending of square-tube specimens [+55, -55, +55, 90], made from the same material as the tubular specimens, with progressive repeated loading. The dimensions of these specimens were: length = 200 mm, side length = 60 mm, and thickness = 1 mm. Two types of bending test were performed: three-point bending and bending on two opposite edges of the square tube (see Fig. 9). The results are presented in Fig. 10.

The small differences found between the experiment and the simulations can be explained by the fact that large displacements were induced during the experiment, whereas a small displacement assumption was made in the simulations. In addition, internal stresses were not accounted for.

All of these tests produced results which were well correlated with the simulations, thus validating the identification process.

6- Conclusion

An identification strategy is proposed, based on the use of a hybrid algorithm and the optimization of experimental tests. This algorithm combines the use of a genetic algorithm with the Levenberg-Marquardt local method. It is based on a statistical approach to the evaluation of the studied domain's topology, once the search domain has been reduced. A basic genetic algorithm is used, with the smallest possible number of adjustable parameters. This is complemented by the use of a statistical approach in the identification strategy, in order to deal with a restricted volume of experimental data. This strategy can be used to reduce the amount of experimental data required for correct parameter identification, thus leading to a reduction in the cost of the testing of materials.

When applied to the identification of the material parameters characterizing the behavior of composite laminates, this strategy is efficient. It could nevertheless be improved: in the present study, the material parameters were identified through the use of a set of experimental measurements. Only the mean values of these measurements were used, and their dispersions were not taken into account. It would however also be possible to use a stochastic identification strategy, in which the objective function is reconstructed by including the experimental measurement errors. This type of strategy would then lead to the evaluation of a set of distributions, rather than a set of deterministic parameters.

**Figures**

Fig 1.

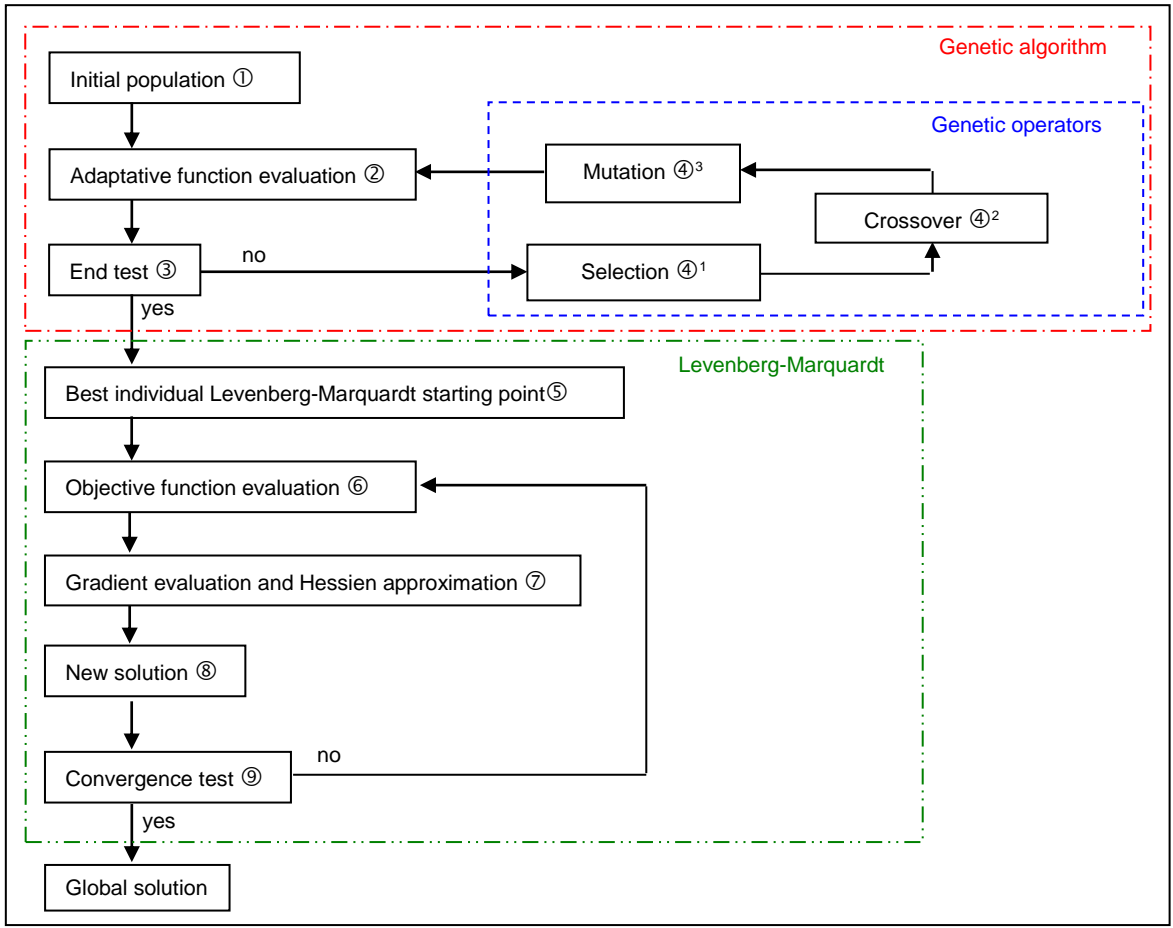

Fig 2.

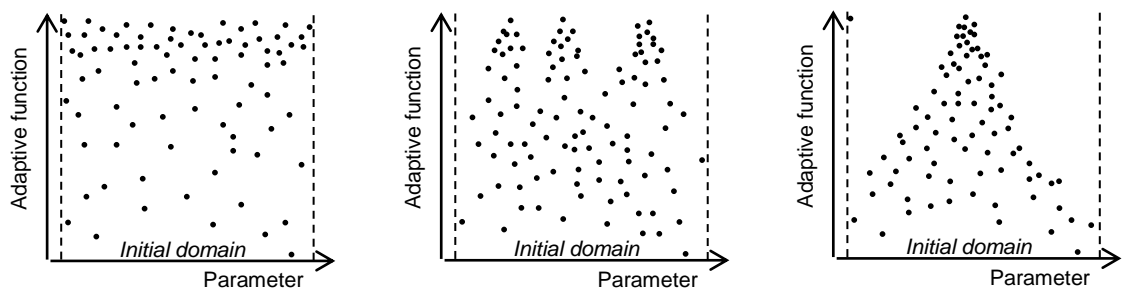

Fig 3.

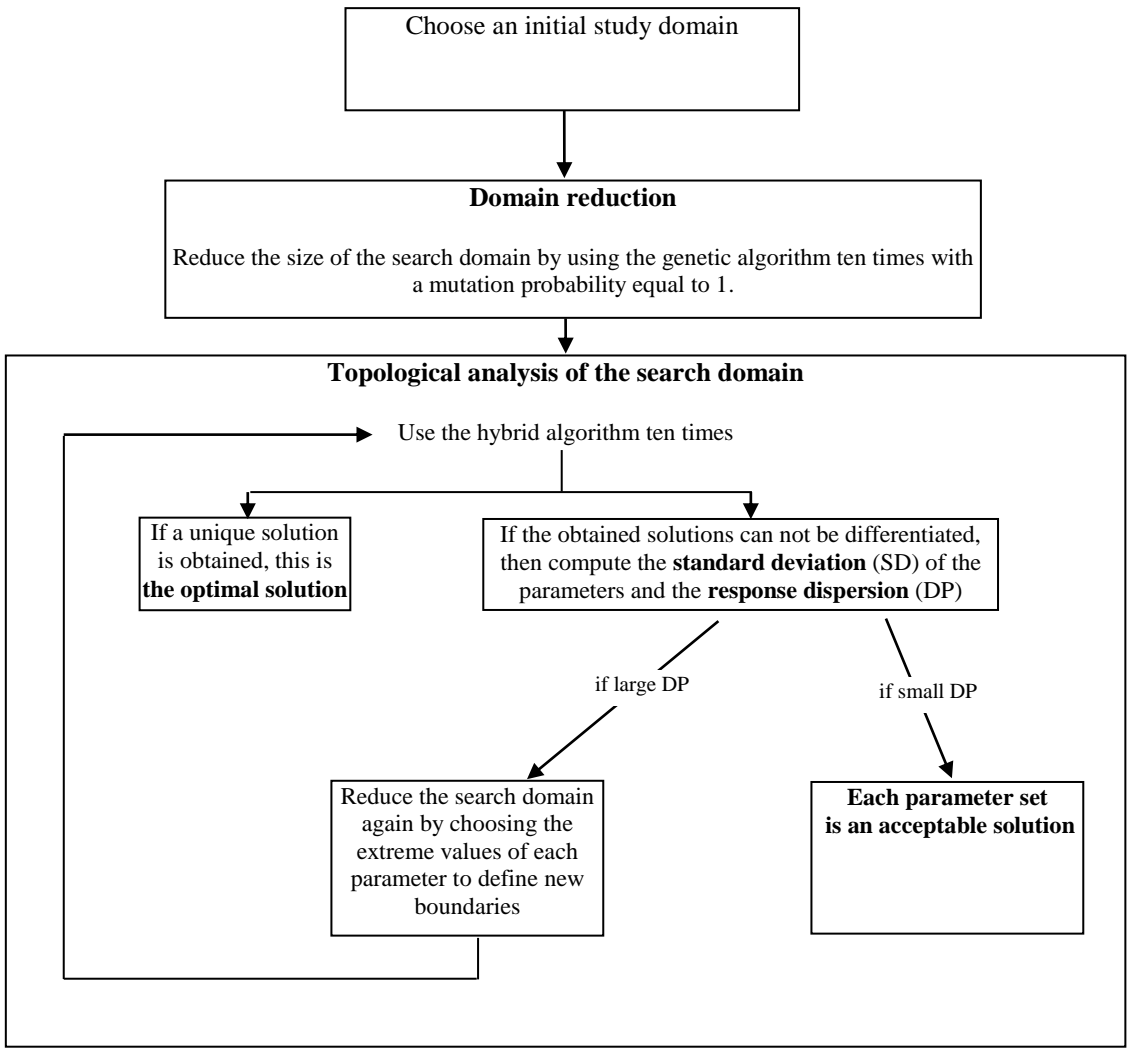

Fig. 4

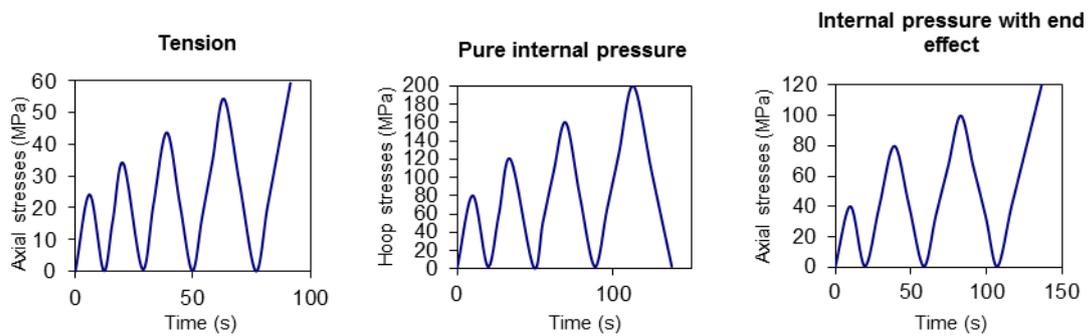

Fig 5.

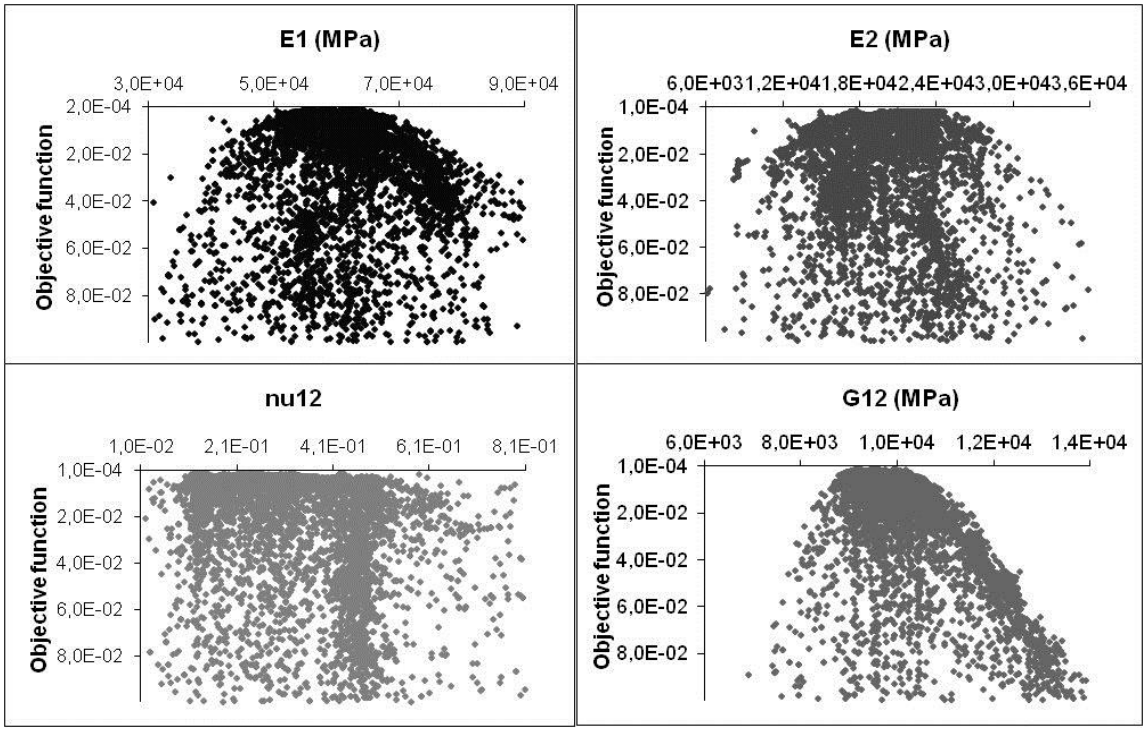

Fig 6.

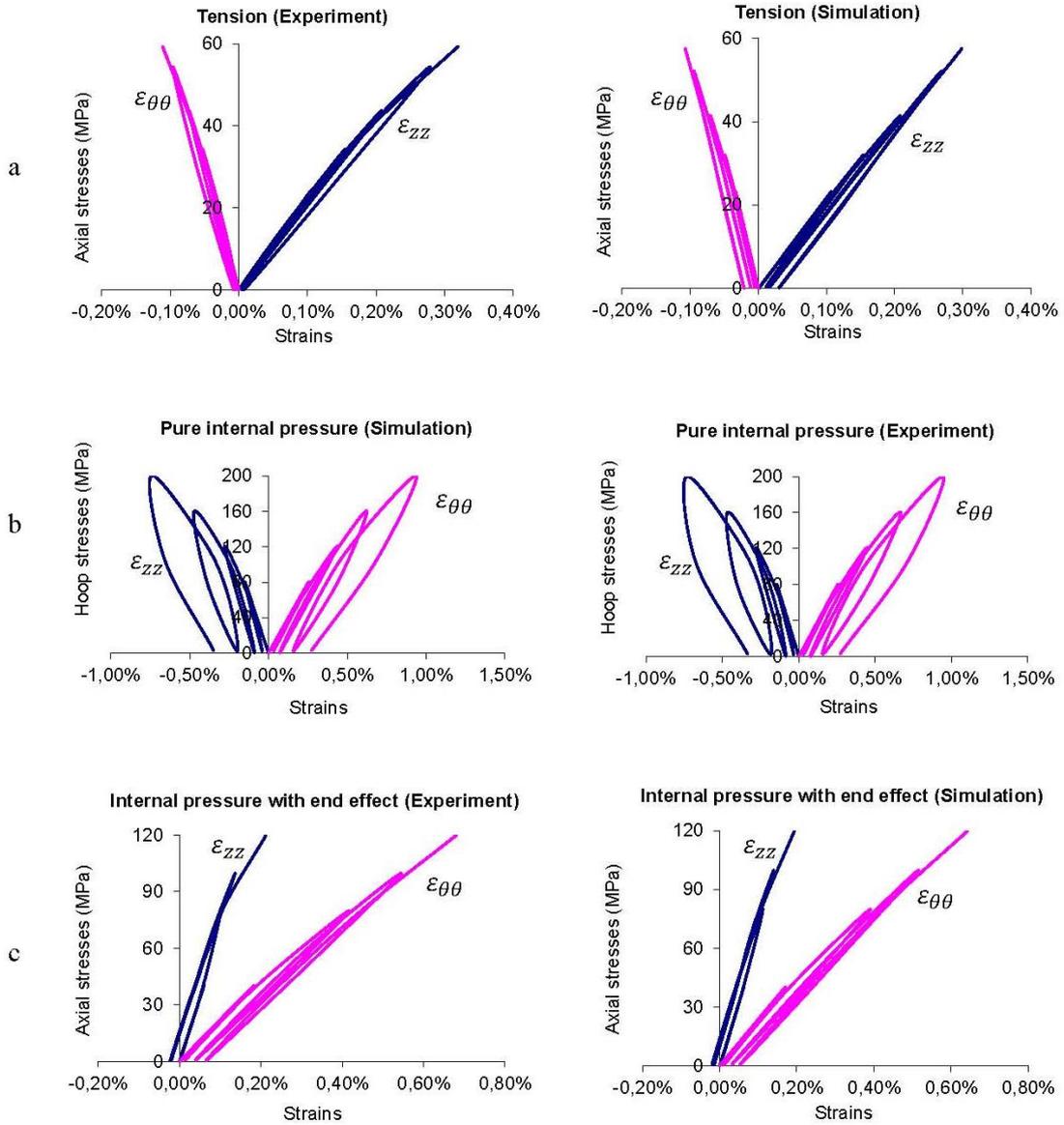

Fig 7.

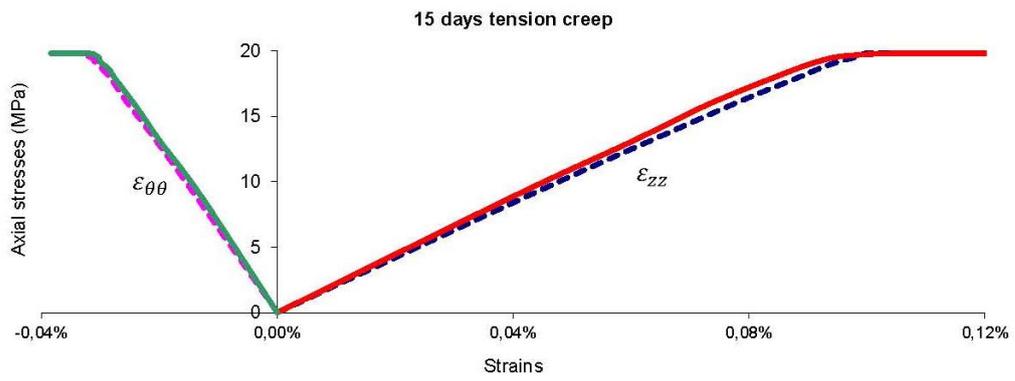

Fig 8.

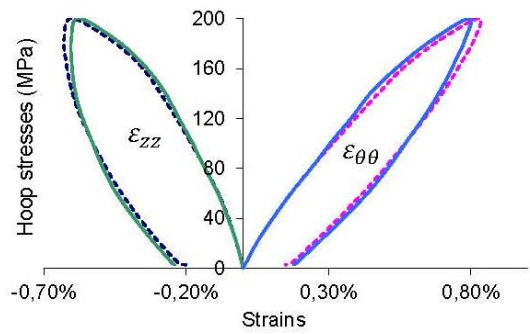

Fig. 9.

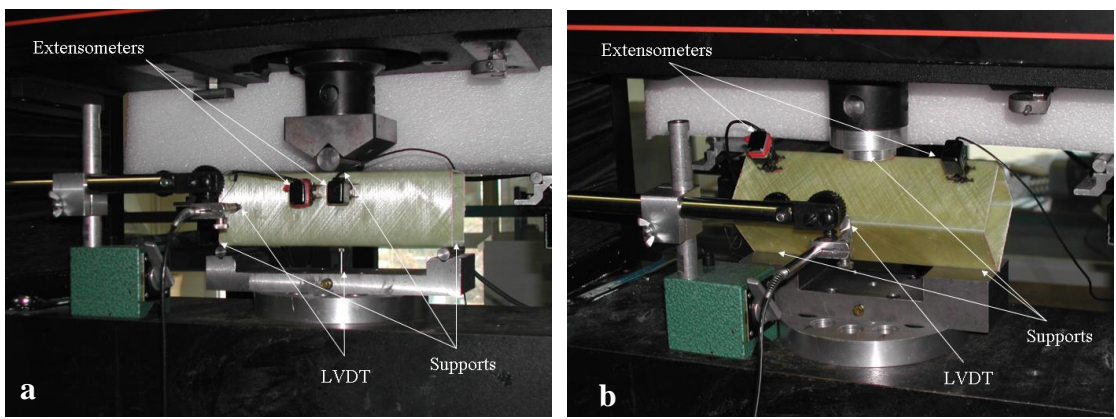

Fig. 10.

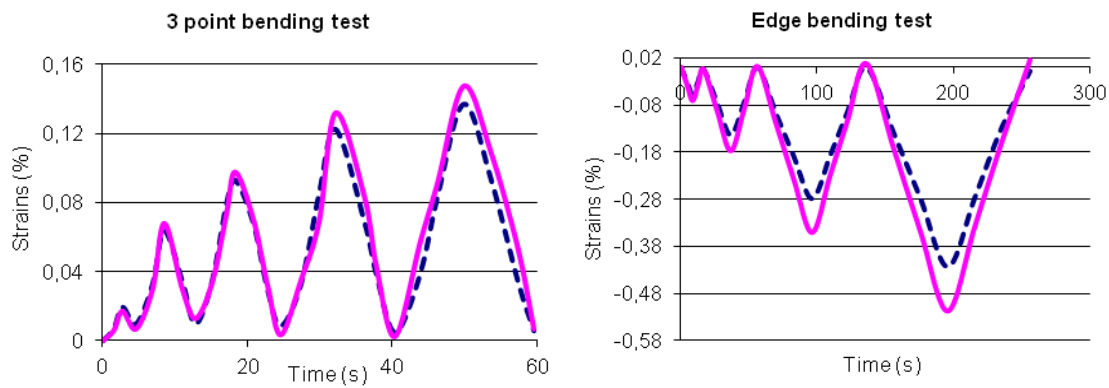

**Figure captions:**

**Fig. 1.** Hybrid algorithm description.

**Fig. 2.** Parameter distributions: (a) uniform distribution (b) distribution with several peaks (c) distribution with one dominant region.

**Fig. 3.** Identification methodology.

**Fig. 4.** Progressive repeated loading tests.

**Fig. 5.** Parameter distributions versus objective function value

**Fig. 6.** Identification tests: comparison between experiments and simulations

**Fig. 7.** Creep under tension: comparison between experiments (dashed line) and simulations (solid line)

**Fig. 8.** Pure internal pressure validation test on a tubular specimen: experimental (dashed line) and simulation (solid line) results

**Fig. 9.** Experimental validation tests: (a) 3 point bending test, (b) edge bending test.

**Fig. 10.** Validation tests: comparison between experiments (dashed line) and simulations (solid line)

**Tables**

Table 1: material parameters of the model described in [5,6]

| Single ply behavior | Parameters |
| --- | --- |
| Elasticity | $E_1$, $E_2$, $\nu_{12}$, $G_{12}$, $G_{23}$ |
| Visco-elasticity | $n_c$, $n_0$, $\beta_{22}$, $\beta_{23}$, $\beta_{44}$ |
| Plasticity | $Z_c$, $\delta_1$, $\delta_2$, $\gamma$ |
| Visco-plasticity | $Z_{vp}$, $\delta_3$, $K$, $\eta$ |
| Damage | $Y_c$, $q$, $p$ |

Table 2: initial and reduced search ranges

| Parameter | Initial range | Reduced range |
| --- | --- | --- |
| $E_1$ (MPa) | [30000; 90000] | [50000; 70000] |
| $E_2$ (MPa) | [8000; 50000] | [14000; 26000] |
| $\nu_{12}$ | [0.019; 0.7] | [0.2; 0.3] |
| $G_{12}$ (MPa) | [3000; 30000] | [8000; 11000] |

Table 3: results of the initial identification

| Draw | $E_1$ (MPa) | $E_2$ (MPa) | $\nu_{12}$ | $G_{12}$ (MPa) | Objective function ($10^{-3}$) |
| --- | --- | --- | --- | --- | --- |
| 1 | 60099 | 21577 | 0.225 | 9450 | 1.7686 |
| 2 | 59216 | 20932 | 0.258 | 9560 | 1.7729 |
| 3 | 65852 | 26617 | 0.212 | 8905 | 1.7697 |
| 4 | 58795 | 20537 | 0.275 | 9600 | 1.7755 |
| 5 | 59371 | 20990 | 0.253 | 9553 | 1.7724 |
| 6 | 60452 | 22085 | 0.215 | 9530 | 1.7708 |
| 7 | 61120 | 22388 | 0.188 | 9395 | 1.7638 |
| 8 | 58429 | 20269 | 0.289 | 9631 | 1.7771 |
| 9 | 59210 | 20910 | 0.252 | 9460 | 1.7802 |
| 10 | 61420 | 20040 | 0.284 | 7640 | 1.7795 |
| Mean | 60396 | 21634 | 0.245 | 9272 | 1.7731 |
| Standard Deviation | 2042 | 1807 | $3.2\ 10^{-2}$ | 578 | $4.9\ 10^{-3}$ |

Table 4: results of the second identification

| Draw | $E_1$ (MPa) | $E_2$ (MPa) | $\nu_{12}$ | $G_{12}$ (MPa) | Objective function ($10^{-2}$) |
|---|---|---|---|---|---|
| 1 | 59850 | 21100 | 0.234 | 9270 | 1.3540 |
| 2 | 59696 | 20988 | 0.230 | 9230 | 1.3577 |
| 3 | 59430 | 21430 | 0.239 | 9200 | 1.3517 |
| 4 | 59900 | 21300 | 0.233 | 9300 | 1.3505 |
| 5 | 59670 | 21360 | 0.238 | 9280 | 1.3512 |
| 6 | 60200 | 21200 | 0.232 | 9220 | 1.3603 |
| 7 | 59930 | 21390 | 0.234 | 9270 | 1.3560 |
| 8 | 60126 | 20928 | 0.231 | 9400 | 1.3576 |
| 9 | 59530 | 21260 | 0.238 | 9200 | 1.3537 |
| 10 | 59950 | 21180 | 0.233 | 9350 | 1.3535 |
| Mean | 59825 | 21213 | 0.234 | 9272 | 1.3544 |
| Standard Deviation | 240 | 160 | $2.6\ 10^{-3}$ | 62 | $3.2\ 10^{-3}$ |

Table 5: identified parameters

| | Parameters | Obtained value | Initial domain |
|---|---|---|---|
| Elasticity | $E_1$ (MPa) | 59900 | [ 35000; 80000 ] |
| | $E_2$ (MPa) | 21300 | [ 10000; 50000 ] |
| | $\nu_{12}$ | 0.233 | [ 0.19; 0.35 ] |
| | $G_{12}$ (MPa) | 9300 | [ 5000; 20000 ] |
| | $G_{23}$ (MPa) | 7700 | / |
| Visco-elasticity | $n_c$ | 14.53 | [ 4; 35 ] |
| | $n_0$ | 13.63 | [ 4; 35 ] |
| | $\beta_{22}$ | 2.51 | [ 0.4; 4 ] |
| | $\beta_{23}$ | 1.35 | [ 0.4; 4 ] |
| | $\beta_{44}$ | 1.44 | [ 0.4; 4 ] |
| Plasticity | $Z_c$ (MPa) | 15.1 | [ 5; 50 ] |
| | $\delta_1$ (MPa) | 24000 | [ 15000; 50000 ] |
| | $\delta_2$ (MPa) | 1926 | [ 1000; 5000 ] |
| | $\gamma$ | 730 | [ 300; 2000 ] |
| Damage | $Y_c$ (MPa) | 0.043 | [ 0.0001; 10 ] |
| | $q$ (MPa) | 1.37 | [ 0.5; 2 ] |
| | $p$ | 0.63 | [ 0.5; 0.99 ] |
| Visco-plasticity | $Z_{vp}$ (MPa) | 15.8 | [ 15; 50 ] |
| | $K$ (MPa) | $2\ 10^{-10}$ | [ $1\ 10^{-13}$; $1\ 10^{-9}$ ] |
| | $\delta_3$ (MPa) | 11300 | [ 2000; 20000 ] |
| | $\eta$ | 5.62 | [ 1; 30 ] |